\documentclass{apa7}
\usepackage[american]{babel}
\usepackage{csquotes}
\usepackage{graphicx}
\usepackage{booktabs}
\usepackage{tabularx}
\usepackage{threeparttable}
\usepackage{setspace}
\usepackage{float}
\usepackage[style=apa,sortcites=true,sorting=nyt,backend=biber]{biblatex}
\DeclareLanguageMapping{american}{american-apa}
\title{Detection of Self-Introductions in Legislative Testimony}
\shorttitle{Self-Introductions in Legislative Testimony}

\author{Sofija Dimitrijevic, Pallavi Das, Kasey Liu, Foaad Khosmood}
\affiliation{California Polytechnic State University}

\leftheader{Dimitrijevic, Das, Liu, \& Khosmood}

\abstract{Self-introductions are common in legislative committee testimonies. Successfully detecting them and extracting the speaker's name is enormously helpful in the task of speaker identification in the context of government meetings. In this paper, we present a pipeline for detection of self-introductions in legislative committee testimony using machine learning. We construct a training dataset from 1.54 million utterances spanning five state legislative sessions, apply a name-matching heuristic to generate automatic labels, and train three classifiers: a decision tree, random forest, and XGBoost to find self-introductions and extract the speaker's name. We construct a feature set combining bag-of-words, positional context, structural signals, introductory phrase indicators, and discourse context features. Among the three classifiers, XGBoost achieves the best performance with an F1 score of 0.9747 and the fewest total errors; adding fine-tuned BERT probability features improves this further. As an extension, we score the full candidate dataset with a fine-tuned BERT classifier and add BERT probability outputs as features. This BERT-augmented XGBoost model improves F1 from 0.9747 to 0.9782 and reduces total test errors from 241 to 207. The primary gain over the decision tree baseline (F1 0.9323) is driven by discourse context features and the boosting ensemble strategy; BERT provides a modest complementary signal. Analysis of false positives reveals that a minority are genuine self-introductions mislabeled due to name inconsistencies in the source data, indicating that measured metrics modestly understate true performance.}


\begin{document}
\singlespacing
\maketitle

\section{Introduction}

\begin{figure}[t]
    \centering
    \includegraphics[width=\columnwidth]{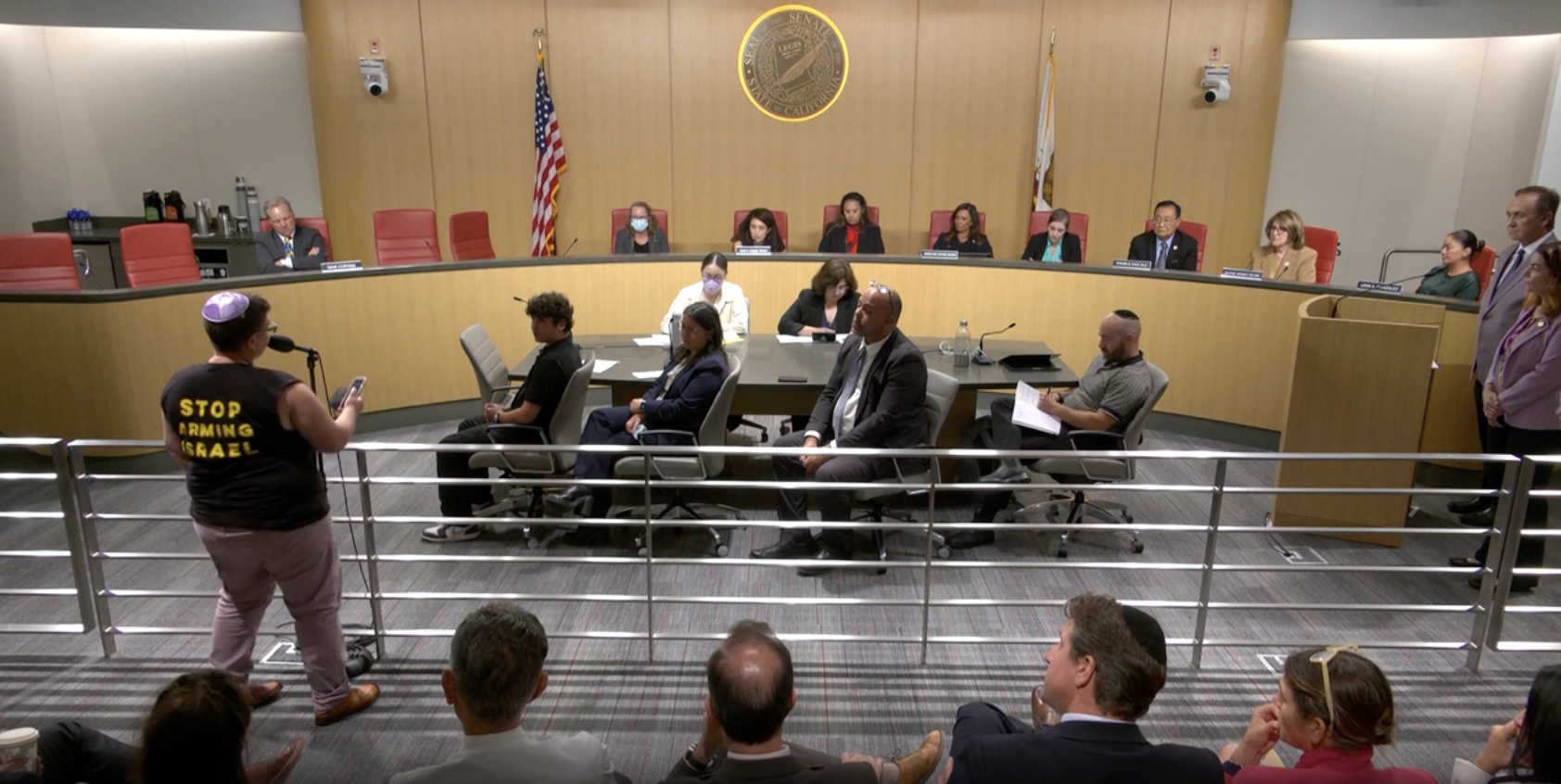}
    \caption{A member of the general public (left, standing) stating a position on AB 715, in Sacramento, California, September 10, 2025.}
    \label{fig:testimony}
\end{figure}

In recent decades, journalistic institutions have seen severe disruption in their processes and economic models, leading to much reduced budgets, capabilities, and coverage, particularly at the local and state levels \parencite{howe2022exploring}. As a result, their essential function in democracy has been diminished, most notably in significant drops in state government reporting due to lack of resources. As reporters cannot be physically present at important government meetings, they increasingly rely on video and transcripts. However, transcripts are often missing or non-comprehensive, lacking essential features such as speaker identification. The Digital Democracy project was created to address this gap using artificial intelligence tools to help reporters with important annotated information, such as speaker identification. It is this latter problem that motivates our work.

Generating comprehensive legislative proceedings is expensive and challenging. Although the basic transcript can be fairly accurately produced using speech-to-text, transcription, and diarization technologies, the speaker identification and annotation parts of the proceedings present difficulties. Specifically, legislative hearings, like many other government committee hearings, are generally attended by two broad groups: known speakers and unknown speakers. The vast majority of the speeches are uttered by known speakers (legislators, staff, other government officials, registered lobbyists), where previous records of their presence allow mature technologies such as voice recognition or face recognition to be used to identify the speakers. Additional clues such as name plates or video annotation may display a legislator's name on screen. However, many unknown speakers have no prior record, no name plates, and no name tags, and the video may not show the correct name at all.

Unknown speakers typically introduce themselves in an opening utterance, meaning that their names are usually mentioned once. To leverage this information, we first identify the speaker as one who fits into the ``unknown'' category of people. Then, we must identify an utterance as the ``self-introductory'' utterance, where someone is first speaking their own name. Lastly, we must extract the correct named entity and present it as a candidate for the identity of the speaker. Subsequent speeches by the same individual can be tagged based on speaker diarization results, which are ordinarily provided at the same time as transcription content. In this way, a substantial portion of a difficult and expensive problem can be addressed with a relatively lightweight solution.

This paper presents a novel pipeline for identifying and extracting speaker names from legislative proceedings by parsing the self-introductory speech of a person. We begin by using the Digital Democracy 2015-2018 corpus of comprehensive proceedings of four state legislatures \parencite{foaad2024digital}. We create an appropriate balanced dataset of utterances labeled with positive and negative self-introduction labels. Then we employ feature engineering to design a classifier that can distinguish between the two classes.

\section{Literature Review}

\subsection{The Digital Democracy Project}

The foundation of this work lies in the Digital Democracy project \parencite{blakeslee2015digital}. The goal of this platform is to promote the transparency and accessibility of state legislative proceedings. Large-scale computational analysis of these proceedings is made possible through the project's aggregation of transcripts, video, and the preparation of metadata from hearings and bill discussions across multiple U.S. state legislatures. \textcite{latner2017measuring} analyzed the platform's ability to study legislative behavior, specifically within California in the 2015-2016 legislative session, through an exploration of participation patterns spanning citizens, legislators, and other parties. This study was one of the first demonstrations that the Digital Democracy corpus could be used for quantitative political analysis, and our work builds directly on and extends this analysis into machine learning classification of utterances using natural language processing (NLP) tools and concepts.

\subsection{Transcription and Speaker Annotation in Legislative Contexts}

\textcite{ruprechter2018} examined the human-assisted transcription pipeline that supports the Digital Democracy platform and identified speaker identification as one of the most time-consuming steps in transcription. Even with advanced professional tools, this process is not fully automated, since speaker identity in transcripts is often inconsistent and noisy. In a subsequent study, \textcite{ruprechter2020} quantified and analyzed tool improvements within the transcription pipeline and found that the inclusion of certain features could increase transcription efficiency by 10\%. Most directly related to our work is \textcite{kauffman2018multimodal}, who used a multimodal approach to address the speaker identification problem within legislative discourse by combining audio, video, and text signals. While their system identifies who is speaking across an entire session, our task is more narrow: given a single utterance, we detect whether the speaker is introducing themselves by name. Their work demonstrates that text alone is not enough for robust speaker identification in this domain, which underscores the value of a dedicated self-introduction detector. This serves as the motivation for our work, since our self-introduction detection and recommendation system can be a useful and lightweight aid for attribution without the need for additional encoding information such as video and audio.

\subsection{NLP and Feature Engineering for Legislative Speech}

Beyond speaker identification, previous work has applied NLP methods to a wide array of legislative analysis tasks. \textcite{grace2023feature} applied feature engineering to legislative hearings to predict stance, organizational affiliation, and engagement levels of public testifiers. Their feature set closely parallels our own approach to feature extraction due to their inclusion of structural and lexical signals extracted from utterances. In addition, their findings informed our inclusion of positional and phrase-indicator features. \textcite{perkonigg2023automatic} applied NLP to recommend relevant bills to statehouse journalists, further demonstrating the breadth of downstream tasks enabled by structured legislative text. For NLP tooling, we rely on spaCy for named entity recognition during candidate selection \parencite{spacy}, a widely adopted library for industrial-strength text processing. While this work focuses on lightweight feature-based classification, recent advances in transformer-based models such as BERT have shown strong performance on related text classification tasks \parencite{devlin2019bert}, and provide a useful comparison point for evaluating whether contextual transformer representations add information beyond sparse lexical and structural features.

\section{Methodology}

\begin{figure*}[p]
    \centering
    \includegraphics[width=\textwidth]{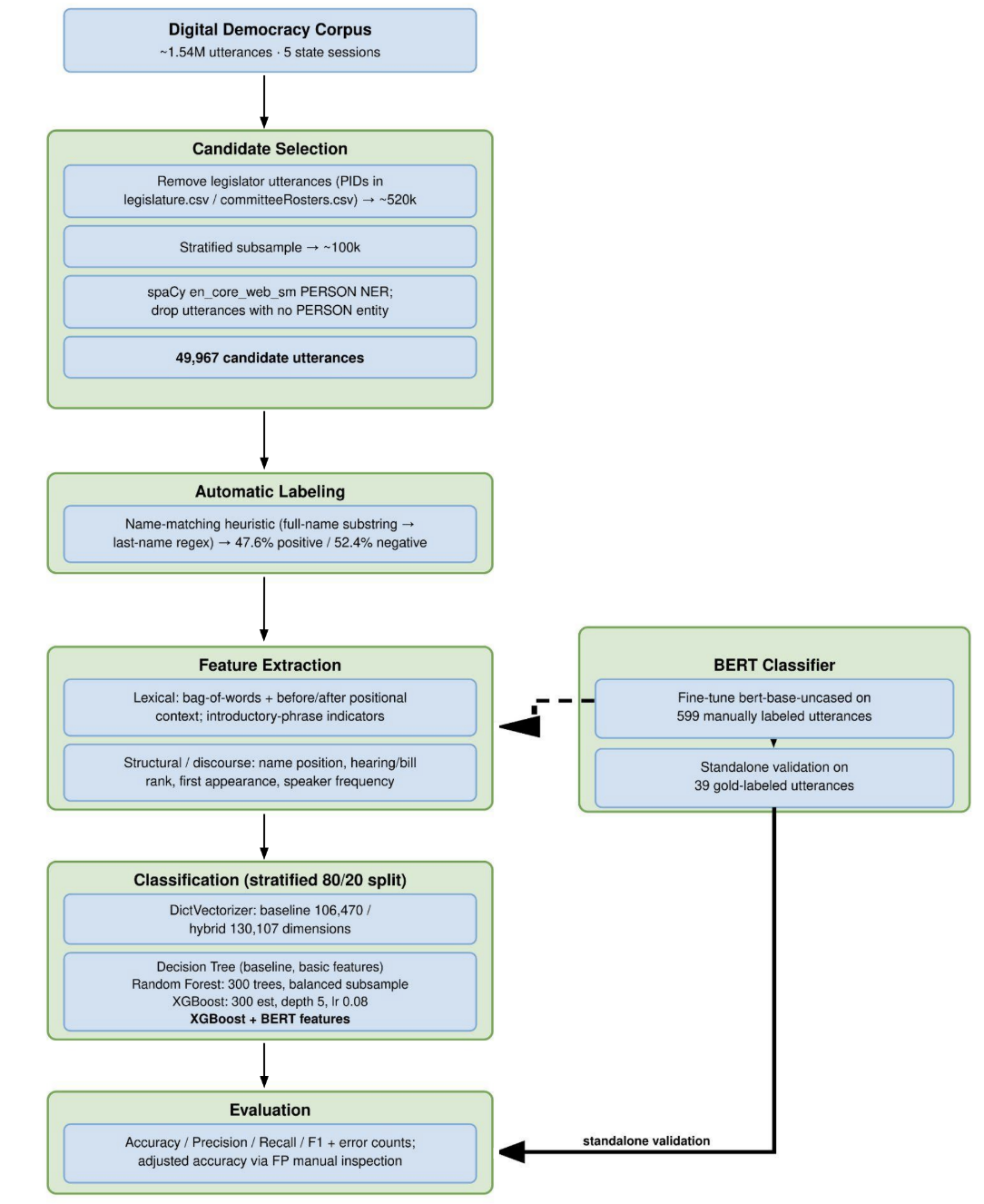}
    \caption{Data Flow Diagram demonstrating the Machine Learning and Model Based Prediction pipeline for self-introduction detection in legislative testimony}
    \label{fig:syspipeline}
\end{figure*}

\subsection{Candidate Selection}

Our training corpus consists of approximately 1.54 million utterances spanning five state legislative sessions (CA 2015-16, CA 2017-18, TX 2017-18, FL 2017-18, NY 2017-18). Since only a small portion of utterances are self-introductions, the corpus is reduced to a more manageable candidate pool. Due to the known institutional role of legislators and their tendency to never introduce themselves, all utterances whose unique personal identifier (PID) appears in \texttt{legislature.csv} or \texttt{committeeRosters.csv} are filtered out. Keeping them would inflate the negative class with structurally distinct speech. This reduces the corpus from 1.54 million to approximately 520,000 non-legislator utterances.

Through a stratified subsample of 100,000 utterances, we ensure that no single state is overrepresented in the candidate pool. To minimize runtime, spaCy's \texttt{en\_core\_web\_sm} model is run over this subsample with named entity recognition (NER) only. Any utterance without at least one PERSON entity is dropped, since a self-introduction must contain a name. Of the 100,000 utterances, 49,967 ($\sim$50\%) remain in the final candidate pool. This final set of 49,967 utterances serves as our gold dataset, upon which all subsequent labeling, feature extraction, and model training are performed.

\subsection{Automatic Labeling}

To label each candidate utterance, a name-matching heuristic is applied. An utterance is labeled as a self-introduction (positive = 1) if the speaker's recorded name appears in the utterance text. After normalizing both name and utterance, the full name is searched as a substring. To prevent partial matching errors, if the full name is not found, the last name is matched using a word-boundary regular expression (\texttt{\textbackslash bLastname\textbackslash b}). Cases where multiple speakers share the same last name are not explicitly handled; such collisions are treated as a known limitation of the heuristic and may contribute to label noise. Utterances with missing or very short last names are labeled negative by default. This produces approximately 23,766 positives (47.6\%) and 26,201 negatives (52.4\%).

\subsection{Feature Extraction}

We extract three classes of features for the feature-engineered models. The first is a bag-of-words representation with additional positional context: tokens before the speaker's detected name are assigned \texttt{before=token} features, and tokens after are assigned \texttt{after=token} features, capturing word-order signals that a flat bag-of-words cannot. The second class captures structural and discourse signals: utterance length, sentence count, normalized name position within the utterance, the utterance's rank within its hearing and bill, whether this is the speaker's first appearance in the hearing, and the speaker's overall utterance frequency. Finally, binary indicator features flag common introductory phrases: ``my name is'', ``representing'', and ``on behalf of''. All features are assembled into a sparse matrix using scikit-learn's \texttt{DictVectorizer}. The decision tree baseline uses basic lexical and structural features (106,470 total dimensions), while the hybrid ensemble adds discourse context features and optionally the BERT probability score (130,107 total dimensions). All models are trained and evaluated on the same stratified 80/20 split: 39,973 training utterances and 9,994 test utterances.

\subsection{Model Selection and Training}

We train and evaluate three classifiers on the same split, chosen to compare a single-tree baseline against two distinct ensemble strategies. A single decision tree provides an interpretable baseline with basic features, serving as a deliberate lower bound for evaluating ensemble and feature gains. Random forest and XGBoost represent two distinct ensemble strategies, bagging and boosting respectively, both trained on the full hybrid feature set. Both ensemble methods are well suited to the sparse, high-dimensional feature space produced by our bag-of-words representation. All three classifiers are implemented using scikit-learn \parencite{scikit-learn}. We train a random forest of 300 trees with balanced subsampling. We train an XGBoost model with 300 estimators, max depth 5, and learning rate 0.08. Unlike random forest, XGBoost trains each tree sequentially on the residual errors of the current ensemble, making it especially effective under class imbalance by concentrating attention on previously misclassified examples.

\subsection{Manual Labeling and BERT Classifier Training}
\label{sec:bert}
The automatic labeling heuristic described relies on name-matching but is subject to noise from name inconsistencies in the source data. To complement this approach and enable transformer-based classification, we constructed a small manually annotated ground-truth dataset. A sample of 599 utterances was drawn from the California 2017-2018 session of the Digital Democracy corpus and annotated by hand. Each utterance was labeled with whether it constituted a self-introduction, producing 116 positive examples (19.4\%) and 483 negative examples (80.6\%).

Before developing the BERT classifier, we explored a regex-based baseline using handwritten patterns such as ``my name is'' and ``I am.'' While this captured straightforward cases, it produced many false positives, such as flagging ``I am against this bill'' as a self-introduction,  and missed non-standard introductions that lacked these exact phrases. This led us to shifting into using a learned classifier capable of capturing contextual patterns.

Self-introduction detection is a context-sensitive task, as whether an utterance constitutes a self-introduction depends on surrounding words and discourse context, not just the presence of a name or a trigger phrase. We therefore fine-tuned \texttt{bert-base-uncased} \parencite{devlin2019bert} for binary 
sequence classification using the Hugging Face Transformers library and PyTorch. The manually labeled dataset was split 80/20 into training and validation sets using stratified sampling to preserve class balance. Each 
utterance was tokenized using the BERT tokenizer, truncated or padded to a maximum length of 256 tokens, and processed in batches of 16. The model was fine-tuned for three epochs using the AdamW optimizer with a learning rate of $5 \times 10^{-5}$. After training, the model and tokenizer were saved for downstream use. The saved classifier was then subsequently used in two ways as a standalone classifier evaluated on a held-out set of 39 manually labeled utterances, and as a feature source for the hybrid ensemble, where the BERT self-introduction probability score was added as an input feature to XGBoost.

See Figure \ref{fig:syspipeline} which illustrates the end-to-end system pipeline describing each stage of the machine learning pipeline.

\section{Results}

\begin{table}[t]
  \begin{threeparttable}
  \caption{Summary Statistics of Cleaned Dataset Post-Preprocessing}
  \label{tab:dataset}
  \begin{tabularx}{\linewidth}{@{}Xr@{}}
  \toprule
  Dataset Feature & Count \\
  \midrule
  Total utterances across corpus & 1,534,371 \\
  Unique legislator PIDs removed & 735 \\
  Non-legislator utterances & 520,736 \\
  Stratified sample size & 100,001 \\
  Candidates after PERSON NER filter & 49,967 \\
  Positive labels (self-intro) & 23,766 (47.6\%) \\
  Negative labels (not intro) & 26,201 (52.4\%) \\
  Training set size & 39,973 \\
  Test set size & 9,994 \\
  Baseline features (Decision Tree) & 106,470 \\
  Hybrid features (XGBoost) & 130,107 \\
  \bottomrule
  \end{tabularx}
  \end{threeparttable}
\end{table}

\begin{table}[t]
  \begin{threeparttable}
  \caption{Model Performance on Test Set}
  \label{tab:performance}
  \begin{tabularx}{\linewidth}{@{}Xcccc@{}}
  \toprule
  Model & Accuracy & Precision & Recall & F1 \\
  \midrule
  Decision Tree & 0.9359 & 0.9363 & 0.9283 & 0.9323 \\
  Random Forest & 0.9535 & 0.9303 & 0.9752 & 0.9522 \\
  XGBoost       & 0.9759 & 0.9726 & 0.9769 & 0.9747 \\
  XGBoost + BERT & 0.9793 & 0.9777 & 0.9788 & 0.9782 \\
  \bottomrule
  \end{tabularx}
  \end{threeparttable}
\end{table}

\begin{table}[t]
  \begin{threeparttable}
  \caption{Error Counts on Test Set}
  \label{tab:errors}
  \begin{tabularx}{\linewidth}{@{}Xccc@{}}
  \toprule
  Model & False Positives & False Negatives & Total Errors \\
  \midrule
  Decision Tree  & 300 & 341 & 641 \\
  Random Forest  & 347 & 118 & 465 \\
  XGBoost        & 131 & 110 & 241 \\
  XGBoost + BERT & 106 & 101 & 207 \\
  \bottomrule
  \end{tabularx}
  \end{threeparttable}
\end{table}

Table~\ref{tab:dataset} summarizes the dataset statistics after preprocessing. As shown in Tables~\ref{tab:performance} and~\ref{tab:errors}, the results confirm our hypothesis: ensemble methods outperform the single decision tree baseline, and XGBoost outperforms random forest across every metric. XGBoost achieves the highest accuracy (0.9759), precision (0.9726), recall (0.9769), and F1 (0.9747), with the fewest total errors (241). This clean ladder going from Decision Tree (F1 0.9323) to Random Forest (0.9522) to XGBoost (0.9747)  holds on the controlled re-run in which all models are evaluated on the same label set, confirming that the gains are real and not an artifact of label-set differences present in earlier analyses. The primary driver of improvement from the decision tree to XGBoost is the combination of the hybrid discourse feature set and the boosting ensemble strategy.

Notably, manual inspection of false positives reveals that some are genuine self-introductions where the speaker's transcript name does not match the name on record (e.g., ``Jose Nunez'' vs.\ ``de Nunez,'' ``Pam Ong'' vs.\ ``Pam Ahlin'').

Qualitative inspection of errors reveals two distinct patterns. Among false positives, label noise is a meaningful but secondary contributor. In a portion of cases, the speaker genuinely introduces themselves, but their transcript name does not match the name on record. The majority are true model errors. For example, ``Jose de Nunez'' says ``my name is Jose Nunez,'' and ``Pam Ahlin'' introduces herself as ``Pam Ong.'' In both cases, the model correctly identifies a self-introduction, but the heuristic labels it negative because the names do not match. To account for this label noise, we manually inspected a sample of false positives and estimated the proportion that were genuine self-introductions mislabeled by the heuristic. Extrapolating this rate across all false positives yields the adjusted accuracies shown in Table~\ref{tab:adjusted}. These findings suggest that the true model performance is modestly higher than the reported figures.

Among false negatives, the largest category consists of utterances where the speaker introduces themselves without an explicit introductory phrase, such as ``my name is.'' For example, ``I'm Jared O'Brien with the Youth Justice Coalition and we strongly support this bill'' and ``Shannon Smith-Crowley, American Association of University Women in California in support'' are both genuine self-introductions that the model misses because they lack the lexical patterns the classifier learned to associate with introductions. See Table~\ref{tab:examples} for additional examples.

\begin{table}[t]
  \begin{threeparttable}
  \caption{Adjusted Accuracy Estimates After False Positive Manual Inspection}
  \label{tab:adjusted}
  \begin{tabularx}{\linewidth}{@{}Xccc@{}}
  \toprule
  Model & Reported & Adjusted & Mislabeled FPs \\
  \midrule
  Decision Tree      & 0.9359 & 0.9432 & 73 / 300 \\
  Random Forest      & 0.9535 & 0.9692 & 157 / 347 \\
  XGBoost            & 0.9759 & 0.9792 & 33 / 131 \\
  XGBoost + BERT     & 0.9793 & 0.9826 & 33 / 106 \\
  \bottomrule
  \end{tabularx}
  \begin{tablenotes}[para,flushleft]
  {\small \textit{Note.} FP-only extrapolation from a reproducible 15\% reviewed sample (seed 42): Decision Tree 11/45, Random Forest 24/53, XGBoost 5/20, XGBoost + BERT 5/16 judged genuine self-introductions. Estimated mislabeled FPs = rate $\times$ total FP; adjusted accuracy credits those back as correct.}
  \end{tablenotes}
  \end{threeparttable}
\end{table}

\subsection{BERT-Augmented XGBoost Experiment}

To evaluate whether contextual transformer features improve performance beyond the hand-engineered feature set, we scored all 49,967 candidate utterances using the fine-tuned BERT self-introduction classifier described in Section~\ref{sec:bert}. For each utterance, the BERT self-introduction probability and binary prediction were added as additional features to the XGBoost model. As shown in Table~\ref{tab:bert}, the BERT-augmented XGBoost achieves an F1 of 0.9782, compared to 0.9747 for XGBoost without BERT features, which is a gain of approximately 0.0035. Total errors decreased from 241 to 207. This suggests that BERT provides useful semantic information but is most effective when combined with the original lexical, positional, and structural features rather than used as a standalone classifier.

The BERT-augmented model reduced both error types compared with the original XGBoost: false positives decreased from 131 to 106 and false negatives from 110 to 101. This indicates that adding the BERT probability improved both precision (0.9726 to 0.9777) and recall (0.9769 to 0.9788), rather than trading one for the other. Standalone BERT, evaluated on 39 manually labeled gold utterances, achieved an accuracy of 0.9487, precision of 0.75, recall of 1.00, and F1 of 0.8571. Note that this evaluation is on a different, smaller set and is not directly comparable to the weak-label test results. The strongest approach is not replacing feature engineering with BERT, but using BERT-derived contextual scores as a complementary signal within the feature-engineered ensemble.

\begin{table*}[t]
  \begin{threeparttable}
  \caption{Example Utterance Errors}
  \label{tab:examples}
  \begin{tabularx}{\linewidth}{@{}Xp{2.5cm}p{2.5cm}@{}}
  \toprule
  Utterance & Speaker & Result \\
  \midrule
  Aton Zitsu here for the Apartment Association California Seven Cities Apartment Association of Orange County, East Bay Rental Housing Association, Nor Cal Rental Property Association, North Valley Property Owners Association and we're also opposed. I would echo what Mr. Moran just spoke to. & Aton Zeitsu & False Positive \\
  Hi, Lupita Cortez Alcal on behalf of the California Student Aid Commission. We are extremely pleased and grateful for the administration's investment in this critical infrastructure. & Lupita Alcala & False Positive \\
  Hello, and thank you for having me today. My name is Lisa Scott, and I've been a provider for 10 years. & Lisa Scott & False Negative \\
  Good morning, Glenn Backes for Drug Policy Alliance. We worked with the Institutes of Justice, ACLU, Assemblymember Hadley, and Senator Mitchell. & Glenn Backes & False Negative \\
  \bottomrule
  \end{tabularx}
  \end{threeparttable}
\end{table*}

\begin{table*}[t]
  \begin{threeparttable}
  \caption{BERT-Augmented Model Comparison on Test Set}
  \label{tab:bert}
  \begin{tabularx}{\linewidth}{@{}Xcccccc@{}}
  \toprule
  Model & Accuracy & Precision & Recall & F1 & FP & FN \\
  \midrule
  BERT standalone (gold labels)* & 0.9487 & 0.7500 & 1.0000 & 0.8571 & 2 & 0 \\
  XGBoost (no BERT features) & 0.9759 & 0.9726 & 0.9769 & 0.9747 & 131 & 110 \\
  XGBoost with BERT features & 0.9793 & 0.9777 & 0.9788 & 0.9782 & 106 & 101 \\
  \bottomrule
  \end{tabularx}
  \begin{tablenotes}[para,flushleft]
  {\small \textit{Note.} Total errors: 2 (BERT standalone), 241 (XGBoost), 207 (XGBoost + BERT). *BERT standalone is evaluated on 39 manually labeled gold utterances and is not directly comparable to the weak-label test set results above.}
  \end{tablenotes}
  \end{threeparttable}
\end{table*}

\section{Discussion}

Speaker identification in legislative transcripts is a hard and expensive problem. Our pipeline of self-introduction detection is only a lightweight solution. Of the machine learning models evaluated, XGBoost achieved the best overall performance across all metrics, with the hybrid discourse feature set driving the majority of the gain over the decision tree baseline. We have shown that high accuracy can be achieved with interpretable textual features, without the need for additional media sources such as audio or video. Our manual inspection of our programmatic label generation suggests that in a portion of false-positive cases our models are correct when the label is wrong, meaning true model performance likely modestly exceeds these measured metrics. Additionally, this indicates that our name-matching heuristic introduces unavoidable noise that would either need to be human-verified or explored in future work. Overall, this work is a useful step in solving the speaker attribution problem within the Digital Democracy project. There are also implications for the broader work of automated transcription pipelines, as well as for the broader mission of the Digital Democracy project: using modern AI tools to close the gap left by declining statehouse journalism and make legislative proceedings more publicly transparent and accessible.

\subsection{Comparing Feature Engineering and Transformer Approaches}

The two approaches explored in this paper represent meaningfully different tradeoffs. The feature-engineered ensemble, trained on weakly labeled data at scale, achieves strong performance (F1 0.9747) and is highly interpretable as individual features such as phrase indicators, positional signals, and discourse context can be directly inspected to understand model predictions. It requires no GPU, runs efficiently at inference time, and scales to tens of thousands of utterances without manual annotation effort. These properties make it well suited for resource-constrained or latency-sensitive deployments, and for teams with energy or infrastructure constraints where transformer-based inference is impractical.

On the other hand, BERT requires only a small manually labeled dataset to fine-tune and captures implicit self-introduction patterns that lexical features miss, such as utterances like ``Glenn Backes for Drug Policy Alliance'' that lack any explicit introductory phrase. However, BERT standalone (F1 0.8571 on 39 gold-labeled utterances) under-performs the feature-engineered ensemble, and introduces significant computational overhead at both training and inference time. The strongest result in this work comes from combining both using the BERT probability score as an additional feature within XGBoost yields F1 0.9782, suggesting that transformer representations and hand-crafted features can be complementary rather than competing. For practitioners facing similar tradeoffs, we recommend the feature-engineering approach when interpretability, speed, or compute efficiency is the priority, and the hybrid approach when a small amount of human-labeled data is available and marginal accuracy gains justify the additional infrastructure cost.

\subsection{Future Work}

There is work to be done to improve the labeling heuristic in the early steps of our pipeline. One such fix could be to implement edit distance matching on the second word of a name, if the first word matches exactly, to catch transcript misspellings of the last name. We could also better handle edge cases within our name-matching heuristic where the record name is ``unknown'' or ``none'' by falling back to matching by first name only. These improvements would reduce false negatives caused by measurement error rather than model error. We also plan to run an unbalanced data experiment, running similar experiments on the full unbalanced dataset to compare against the current stratified balanced sample and reveal how our models perform under a real-world class distribution, where self-introductions are rare. Future work should fine-tune and evaluate transformer models on a larger manually labeled dataset. Although BERT improved the ensemble methods when used as an additional feature, the current evaluation still relies heavily on weak labels, so a larger human-verified benchmark would better measure true performance.

\section*{Acknowledgments}

[ redacted ]

\printbibliography

\end{document}